\documentclass[sigconf, authorversion]{acmart}

\makeatletter
\renewcommand\@formatdoi[1]{\ignorespaces}
\makeatother
\renewcommand\footnotetextcopyrightpermission[1]{} 
\usepackage{balance}

\usepackage{amsmath,amsfonts}
\usepackage{graphicx}
\usepackage{balance} 
\usepackage{subcaption}
\usepackage{stackengine}
\usepackage{mwe}

 \newlength{\tempdima}
 \newcommand{\rowname}[1]
 {\rotatebox{90}{\makebox[\tempdima][c]{\textbf{#1}}}}

\begin{document}

\title{\texttt{sat2pc}: Estimating Point Cloud of Building Roofs from 2D Satellite Images}

\author{Yoones Rezaei}
\email{yor10@pitt.edu}
\affiliation{
  \institution{University of Pittsburgh, Pittsburgh}
}
\author{Stephen Lee}
\email{stephen.lee@pitt.edu}
\affiliation{
  \institution{University of Pittsburgh, Pittsburgh}
}

\begin{abstract}
Three-dimensional (3D) urban models have gained interest because of their applications in many use-cases such as urban planning and virtual reality. However, generating these 3D representations requires LiDAR data, which are not always readily available. Thus, the applicability of automated 3D model generation algorithms is limited to a few locations. In this paper, we propose \texttt{sat2pc}, a deep learning architecture that predicts the point cloud of a building roof from a single 2D satellite image. Our architecture combines Chamfer distance and EMD loss,  resulting in better 2D to 3D performance. We extensively evaluate our model and perform ablation studies on a building roof dataset. Our results show that \texttt{sat2pc} was able to outperform existing baselines by at least 18.6\%. Further, we show that the predicted point cloud captures more detail and geometric characteristics than other baselines.

\end{abstract}

\maketitle
\pagestyle{plain}

\section{Introduction}

There is a growing need for three-dimensional (3D) representations of urban models, which play an important role in many useful applications such as urban monitoring, utility services, transportation and virtual reality. Much of the current research to date primarily relies on aerial-based lidar point cloud data to extract accurate and fine-grained 3D representations of urban models. As such, there has been considerable effort towards accelerating the collection of lidar point cloud data, commonly used to develop 3D representations\cite{3D_elevation_program}. Existing programs spend over a hundred million dollars annually to collect 3D point clouds\cite{3D_elevation_program_cost}.   
  
Unfortunately, it is unlikely that the costs of collecting aerial-based lidar data to decline in the near future. Aerial-based lidar data involves flying airplanes and drones with a laser scanning system that directly measures three-dimensional point coordinates of the underlying surfaces. While they provide accurate depth information, it limits the application of existing state-of-the-art techniques to only those locations where lidar data is available. We note that large parts of the world still have no lidar coverage. As such, it makes it challenging to benefit from any advances in technologies that use lidar data to generate urban models. Another limitation of lidar data is that they may be outdated. Because of the high costs, lidar data are not frequently updated, and many data are over a decade old\cite{Opentopography}. Thus, any recent urban developments will be left out, especially in fast-growing cities that are constantly changing.

To solve this issue, researchers have attempted to generate 3D models using one or more 2D images. Recent advancements in deep neural architecture have rapidly improved the reconstruction of 3D models using single or multiple images, learning the 3D geometrical properties\cite{fan2017point, nguyen2019graphx, wen2019pixel2mesh++}. However, most existing techniques use data derived from clean 3D CAD models to train and reconstruct point cloud representations, making it easy to learn shape models\cite{shapenet2015}. Unfortunately, point cloud data from lidar sensors are inherently noisy, irregular, and non-uniform. Occlusions from nearby trees and objects may distort the measurements. Thus, reconstructing a 3D point cloud of urban models from images will need to account for noise inherent in aerial-based lidar data.

In this paper, we propose \texttt{sat2pc} that predicts 3D point cloud of building roofs from single 2D satellite images. Our key hypothesis is that advances in deep learning-based techniques have made it feasible to extract and reconstruct 3D aspects of an object from 2D images. Since satellite images are widely and freely available through mapping services such as Google or Bing Maps, the ability to derive 3D point clouds can then automate the construction of urban models in a data-driven and scalable manner. Moreover, satellite images provide rich features and edges that enable the identification of building outlines and roof planes. Given this information, humans can, to an extent, reconstruct 3D shape of the roof and different planes. Our key research question involves addressing whether it is feasible to reconstruct 3D roof shapes by generating a point cloud representation. We note that in addition to satellite imagery, high-resolution lidar data is also available for download using the 3D Elevation program\cite{3D_elevation_program}. Consequently, we can develop data-driven models for reconstructing point clouds using satellite images, and thus, such an approach can benefit from the decades of research on fusing point clouds with imagery to develop urban models.

While there has been recent work on estimating 3D point cloud from single 2D images\cite{nguyen2019graphx, fan2017point}, there are important differences between existing works and our proposed research. 
First, existing 2D to 3D techniques produce either density-aware point cloud or detail-aware that pays attention to the overall structure but are irregular with imbalance point distribution. On the other hand, our approach captures both detail-aware and the density of the points. 

Second, much of the existing work focuses on reconstructing common objects. In contrast, our approach focuses on extracting the geometrical features of roofs and predicting point cloud representations. 
There have also been studies on estimating height from single 2D images\cite{liu2020im2elevation, ghamisi2018img2dsm, srivastava2017joint}. However,  these techniques are orthogonal to our proposed work, as we primarily focus on reconstructing roof shapes. Thus, we can combine these techniques to create building models at scale.

We present \texttt{sat2pc} that addresses the problem of extracting geometric details of the roofs. In doing so, we make the following contributions:

\noindent
\textbf{\texttt{sat2pc} Framework:} 
We present a novel architecture that predicts building point clouds using a single 2D image. Our technique can capture the fine-grained detail and reconstruct a rich representation of the 3D roof. Moreover, our architecture can leverage the strengths of Chamfer distance and Earth Mover's Distance (EMD) loss to achieve a detail-aware and density-aware point cloud.

\noindent
\textbf{Building Point Cloud Dataset:} 
We created a building dataset that consists of 343 satellite images and their corresponding lidar-based point cloud data. The images and lidar data are annotated with building outline and point cloud classification labels. 
We manually inspected the data and fine-tuned the projections to align the lidar and image data. 

\noindent
\textbf{Evaluation:} 
We extensively evaluated our technique and performed ablation studies to show its efficacy in predicting point cloud representations. Our analysis shows that \texttt{sat2pc} achieves 18.6\% improvement over the best baseline model. Further, our results indicate that the point cloud is density and detail-aware. We also used the existing planar extraction algorithm to segment planar roof segments from the predicted point cloud. The results indicate that \texttt{sat2pc} can reconstruct the geometric aspect (e.g., planes) in the point cloud. 

\section{Background}
\label{sec:background}
\subsection{Building Data and Reconstruction} 

Three-dimensional (3D) building models can be generated from various data sources. For example, aerial images (e.g., satellites) have high-resolution (less than a foot per pixel) data that captures detailed close-up characteristics of buildings and other urban features. They provide building outline and geometrical aspects (e.g., orientation and shape). There are also aerial-based lidar data with accurate depth information of urban objects, ideal for urban modeling. Airborne lidar provides irregular and unordered 3D point measurements and requires preprocessing to create a digital elevation model (DEM). Both data are typically incomplete and each data sources have some parts that are in greater detail than others. For instance, aerial images lack depth information but provide texture and rich image details missing in lidar data.
 
Prior approaches focused mainly on aerial 2D images to model cities at large-scale\cite{duan2016towards, mahmud2020boundary}. These techniques typically segment building outlines using vision-based algorithms to produce polygonized footprints of the building\cite{alidoost20192d}. In addition, building heights can be estimated from multiple or single images to generate the final 3D building model. However, most models often lack detail and assume flat roofs for all buildings. Furthermore, they usually require multi-view images of a building to reconstruct a rich model and are thus not scalable. It is possible to add some details into the model by forming a hypothesis of roof shapes and mapping them to a standard architectural form of the roof\cite{alidoost20192d}. However, this lacks realism and fails to capture the variety that may occur in the real world. As such, most tools and techniques today are designed to use both lidar and high-resolution images to obtain accurate models. 

Unfortunately, lidar data is expensive to collect and not available for large parts of the world. 

\subsection{Inferring 3D Shapes from Single 2D Images}
In recent years there have been several efforts to predict the 3D shapes of common objects (e.g., ShapeNet\cite{shapenet2015}) and urban environments (e.g.,  cars in Kitti dataset\cite{Geiger2013IJRR}). However, little focus has been on estimating fine-grained details of 3D building models using single 2D images. Recently, \cite{liu2020im2elevation, mahmud2020boundary, ghamisi2018img2dsm, li20213d} take advantage of deep neural network-based approaches such as GAN, CNNs and, encoder decoders to estimate a  \textbf{D}igital  \textbf{S}urface  \textbf{M}ap (DSM) or height of buildings using a single aerial image.  But they do not focus on estimating the fine-level details of the 3D building. In contrast, our work focus on reconstructing the 3D point cloud representation of a building from 2D satellite images. Note that several techniques fuse lidar point cloud and satellite imagery to create urban models, and thus, can benefit from having a 3D point cloud representation.  

\subsection{Point Cloud Similarity Metrics}
\label{sub_sec:pc_metrics}
\subsubsection{\textbf{C}hamfer \textbf{D}istrance:} 
CD is one of the two broadly adopted metrics for calculating the similarity of two point clouds. We define CD as:\\
\begin{equation}
    \mathcal{L}_{CD}(P_1, P_2) = \frac{1}{|P_1|}\sum_{x \in P_1} \min_{y \in P_2} \| x-y \|_{2}^{2} + \frac{1}{|P_2|}\sum_{y \in P_2} \min_{x \in P_1} \| x-y \|_{2}^{2}
\label{eq:cd}
\end{equation}

where, $P_1$ and $P_2$ are two point clouds. Each point finds the closest point in the other point set (and vice-versa) and calculates the average pair-wise point-level distance. CD lacks density awareness but is stricter with capturing details~\cite{wu2021density}. This causes the model to cluster most points in regions with high point density (e.g., the object's center).

\subsubsection{\textbf{E}arth \textbf{M}over's \textbf{D}istance:} 
EMD is another broadly used metric for calculating similarity between two point clouds and is defined as:\\
\begin{equation}
    \mathcal{L}_{EMD}(P_1, P_2) = \min_{\psi:P_1 \rightarrow P_2} \frac{1}{|P_1|}\sum_{x \in P_1} \|x-\psi(x)\|_{2}
\label{eq:emd}
\end{equation}

where $\psi$ is a bijection function from $P_1$ to $P_2$ where EMD finds the $\psi$ with the lowest sum of pair-wise distances and returns the average. Since finding a bijection function with the lowest cost is a computationally expensive task, in practice, an approximation version of EMD is often used \cite{fan2017point, liu2020morphing}. Using an approximation approach also relaxes the limitation of using two point clouds with different sizes.
As a result, EMD is sensitive to distribution changes but is less strict with detailed structures~\cite{wu2021density, fan2017point}.

\section{\texttt{sat2pc} Design}
\label{sec:systemdesign}

\begin{figure*}[t]
\includegraphics[width=.8\linewidth]{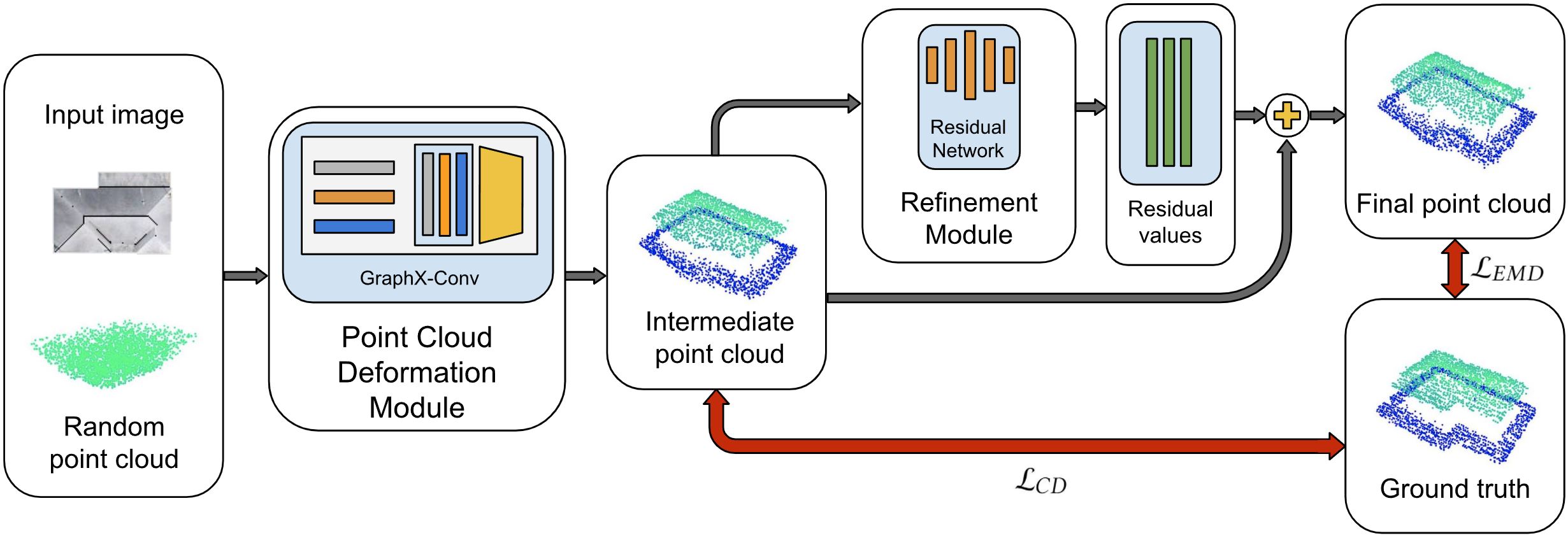}

\caption{\texttt{sat2pc}  architecture consists of two modules. The PCD module takes as input an image and a random point cloud and generates the intermediate point cloud. Next, the refinement module takes the intermediate point cloud and generates the point-wise residuals. These residual values are added to the intermediate point cloud to generate the final output.}
\label{fig:framework}
\end{figure*}

Our primary goal is to produce a 3D point cloud from a single 2D satellite image. Specifically, given an image of size m$\times$n and point sample size $N$, we aim to generate a dense point cloud set $P=\{p_i\}_{i=1}^N$ of size N points. Figure~\ref{fig:framework} depicts our proposed architecture. The \texttt{sat2pc} design has two key components ---  a \textbf{P}oint \textbf{C}loud \textbf{D}eformation (PCD) network and a refinement network. As illustrated in the figure, the PCD network takes input an image and a randomly generated point cloud of size $N$. It is responsible for producing an intermediate point cloud ($P'$) that captures the structural aspects of the input image. The refinement network takes as input the intermediate point cloud $P'$ and fine-tunes them to generate the final point cloud. Our \texttt{sat2pc} architecture is able to enforce different loss functions, Chamfer and EMD loss, to predict the 3D shape of the roof.

\subsection{Point Cloud Deformation Module}
\label{subsec:pcd}
Our input satellite image consists of planar roof segments that make up the overall structure of the building. In this stage, our objective is to extract the geometric features from the satellite image to produce the point cloud. As such, for our initial feature extraction and 3D point cloud generation, we use the GraphX-Conv architecture proposed in \cite{nguyen2019graphx}. The proposed GraphX considers the spatial correlation between points and encodes point-specific and global features of the image onto a random point cloud, which helps reconstruct the final point cloud. Below, we describe the key aspects of GraphX and refer the readers to the original paper~\cite{nguyen2019graphx}.

GraphX-Conv consists of three main steps. First, it has an image encoder to obtain a multi-scale representation of the input image. GraphX uses a similar VGG architecture as described in \cite{nguyen2019graphx}. 
The next step is feature blending, where the point-specific and global shape information are concatenated with an initial randomly generated point cloud. Point-specific shape information is extracted by projecting the initial random points onto the extracted feature maps from the image encoder. And, global shape information is extracted using \textbf{Ada}ptive \textbf{I}nstance \textbf{N}ormalization(AdaIN) \cite{huang2017arbitrary}, a style transfer method which aligns the mean and variance of a sample to a target style. The AdaIN helps to transfer the style of extracted features from the image encoder to the initial point cloud and encode the global shape information. The final step is a deformation that uses a ResGraphX module that generates a point cloud representation of the input image by using the features vector created in the previous step. ResGraphX is a deformation module inspired by X-Convolution \cite{li2018pointcnn} and graph-convolution \cite{kipf2016semi} which has a lower computational cost than X-Convolution and is more flexible than graph-convolution.

The size of the initial random point cloud dictates the final size of the reconstructed point cloud. Since the deformation module deforms the input point cloud, the predicted point cloud retains the same number of points. Although there is no limitation on the number of points of the input cloud, we observed that a large point cloud size could significantly slow down the training process.

\subsection{Refinement Module}
The predicted point cloud from PCD forms an intermediate point cloud in our network and lacks fine-grain structural characteristics of the roof. Although it captures the overall structure, a lack of density-awareness leads to irregular and non-uniform point clouds. Since we are interested in retaining the overall shape of the roof, the characteristics of the individual planar roof segments must be retained in the final output. That is, point cloud points should be distributed uniformly on the same plane for a plane extraction algorithm to detect them as a part of the same plane. 
If retained, off-the-shelf planar extraction can be used to identify and extract these planar segments to evaluate the geometry of the 3D point cloud. To add these fine-grain structural details, we use a similar residual network as suggested by \cite{liu2020morphing}. The residual network takes the point cloud predicted by the PCD module and predicts the point-wise residual. This network consists of seven 1D convolutional layers that generates the a three channel residual for each point(one for each dimension of the point). 
The final point cloud results from adding these residuals to the PCD's output. We use a similar residual network to \cite{liu2020morphing}. However, in our case, the residual network only receives an N$\times$3 point cloud and predicts an N$\times$3 residual vector, where $N$ is the size of the input point cloud.

\subsection{Loss Function}
As described in section \ref{sub_sec:pc_metrics}, Chamfer distance and EMD are the two most popular metrics for measuring similarity between two point clouds. However, each one of these metrics has its strength and shortcomings. For example, Chamfer distance is more detail-oriented but lacks density awareness, whereas EMD is sensitive to density while ignoring local details \cite{wu2021density}. These issues lead us to use both metrics but at different stages of our architecture. This approach helps us to get the best of two worlds and have a detailed point cloud, using Chamfer distance, while utilizing the limited number of available points by having a density-aware metric like EMD. 

Since PCD has direct access to the input image, it can extract geometric features from the satellite image and incorporate more details in the predicted point cloud. As such, we find that having Chamfer distance to calculate the loss for PCD's output results in obtaining better shape information. It ensures the predicted point cloud is detailed and captures the overall shape information. However, as we show in section \ref{sec:experiments}, Chamfer loss clusters point towards a specific part of the predicted point cloud, reducing the overall quality of the 3D shape. The point cloud shape distribution is also irregular and non-uniform, making it difficult for the plane extraction algorithm to extract planar roof segments. 
 On the other hand, the refinement network produces residual values for each point in the point cloud. We use this network to refine the density and ensure uniformity in the predicted point cloud. Thus, we use EMD loss that is density-aware and achieves point distribution similar to the ground truth.We combine the Chamfer and EMD loss, an use $\alpha$ parameter to control the ratio of the different losses. Our final loss is defined as follows.

\begin{equation}
    \mathcal{L}(X_{inter}, X_{final}, Y) = \mathcal{L}_{CD}(X_{inter}, Y) + \alpha \mathcal{L}_{EMD}(Y, X_{final})
\end{equation}

where $X_{inter}$ is the intermediate prediction from PCD module. $X_{final}$ is the final prediction from the refinement module, $Y$ is the ground truth point cloud. And, $\mathcal{L}_{CD}$ and $\mathcal{L}_{EMD}$ are Chamfer and EMD loss, respectively. Our combined loss function makes the predicted point cloud have both the details and density-awareness of Chamfer and EMD loss, respectively.

\section{Evaluation Methodology}
\label{sec:evaluation_methodology}

In this section, we describe our dataset, metrics, and experimental setup to evaluate our proposed approach. 

\subsection{Dataset Collection and Description}

\begin{table}[t]
\footnotesize
\begin{tabular}{c|c|c|c|c|c}
\toprule

\textbf{Dataset}                                                                 & \textbf{Ours}                                             & \textbf{\begin{tabular}[c]{@{}c@{}}DALES\\ \cite{varney2020dales}\end{tabular}}                                                                            & \textbf{\begin{tabular}[c]{@{}c@{}}DFC\\ \cite{c6tm-vw12-19}\end{tabular}}                                                                       & \textbf{\begin{tabular}[c]{@{}c@{}}SpaceNet 2\\ \cite{spacenet2_data}\end{tabular}} & \textbf{\begin{tabular}[c]{@{}c@{}}Vaihingen\\ \cite{vaihingen_data}\end{tabular}} \\ \midrule
\textbf{LiDAR}                                                                   & Yes                                                       & Yes                                                                                         & Yes                                                                                & No                                       & Yes                                      \\ \hline
\textbf{\begin{tabular}[c]{@{}c@{}}Satellite\\ Image\end{tabular}}               & Yes                                                       & No                                                                                          & Yes                                                                                & Yes                                      & Yes                                     \\ \hline
\textbf{\begin{tabular}[c]{@{}c@{}}Image\\ Building Outline\end{tabular}}        & Yes                                                       & No                                                                                          & Yes                                                                                & Yes                                      & Yes                                     \\ \hline
\textbf{\begin{tabular}[c]{@{}c@{}}Lidar Building\\ Classification\end{tabular}} & \begin{tabular}[c]{@{}c@{}}Yes\end{tabular} & Yes                                                                                         & Yes                                                                                & No                                       & No                                      \\  \hline
\textbf{\begin{tabular}[c]{@{}c@{}}Pre-Processed\\ Building Data\end{tabular}} & \begin{tabular}[c]{@{}c@{}}Yes\end{tabular} & No                                                                                         & No                                                                                & No                                       & No                                      \\ \bottomrule
\end{tabular}
\caption{Comparison of our dataset with existing datasets.}
\label{tb:datasets}
\end{table}
While there are existing datasets that provide point cloud\cite{c6tm-vw12-19, vaihingen_data} and satellite images, these datasets have many limitations. For example, DFC dataset's satellite images does not have high enough resolution. Similarly, most high-resolution data are not large-scale or have enough building variety. These datasets consist of mostly high rises that have flat roofs. Thus, we created a high-resolution building dataset of satellite images and lidar information from existing mapping APIs to evaluate our approach. Table~\ref{tb:datasets}  highlights the key differences between our dataset and available datasets. 

Our data collection process is as follows. 
We collected the lidar data \cite{FL_dense, FL_sparse} from OpenTopography website\cite{Opentopography}, which contains high-resolution data for various cities in the US. Next, we downloaded the building outline using the location information in lidar data. We used the OpenStreetMap API that provides the building outline of all homes within a region. Next, we used this outline to extract the corresponding point cloud from the lidar data. In addition,  we used the MapBox API to download the corresponding satellite image for each building using the location coordinates.

While collecting and mapping the lidar and satellite images from existing APIs seems straightforward, we faced many challenges in creating the dataset. We learned from experience that the lidar data sometimes did not match the satellite images. There were instances when the 3D building roof structures had a different shape than their 2D counterpart, indicating either the lidar or satellite data was outdated. As such, we meticulously removed these instances from our dataset. Similarly, we also removed images where the buildings were entirely occluded by nearby trees. In addition, we found instances where the lidar point clouds were noisy --- points had either very high or low values and indicated an error during data collection. To fix this issue involved manually plotting and inspecting each building from multiple angles, which proved to be time-consuming.  In some cases, when the projections were transformed from lidar coordinates (in latitude and longitude) to image coordinates (in x-y coordinates), the building outline did not align well. As a result, we often had to fine-tune the values by adjusting the offset to get accurate mappings between pixel coordinates of images and points in the point cloud.

Our \texttt{sat2pc} roof dataset consists of 343 samples from Dunedin, Florida. The dataset covers different roof types such as flat, gable, heap, and combination of them. Each sample consists of a 224x224 satellite image, a point cloud representing the building, and the building and roof planes outlines. The code for extracting and fine-tuning lidar data and the dataset will be made available to enable future research in this area. 
    
\subsection{Metrics}
\label{subsec:metrics}
We note that standard metrics such as Chamfer distance and EMD are not ideal for capturing the geometric characteristics of the roof. While they measure the general structure and density discrepancy, they do not reveal whether the planar structures of the building roof were reconstructed in the predicted model. For instance, any outlier points may not significantly impact the Chamfer loss but may change the roof's building outline and planar structures. Thus, we use the metrics proposed in \cite{awrangjeb2014automatic} that are commonly used to capture the characteristics of a 3D building roof. These metrics are as follows.

\noindent
\textbf{Completeness} calculates the proportion of ground truth's roof planes that are present in the prediction and defined as:

\begin{equation}
    Cm = \frac{TP}{TP + FN}
\label{eq:cm}
\end{equation}
where $TP$ is the number of planes that are present in both ground truth and prediction point clouds. A plane from prediction is true positive when it's matched with a plane from the ground truth. Plane $Pl_i$ from prediction is a match for plane $G_j$ from ground truth if, their IoU is the largest, and their intersection is larger than or equal to 40\% of the area of $G_j$. And, $FN$ is the number of planes that are present in ground truth but missing in the predicted point cloud. 

\noindent
\textbf{Correctness} measures the proportion of prediction's roof planes that are present in the ground truth and is defined as:

\begin{equation}
    Cr = \frac{TP}{TP + FP}
\label{eq:cr}
\end{equation}

where $FP$ is the number of planes that are present in the prediction's point cloud but do not exist in the ground truth. 

\noindent
\textbf{Quality} measures the similarity between the ground truth and predicted points and is defined as:

\begin{equation}
    Q = \frac{TP}{TP + FP + FN}
\label{eq:cr}
\end{equation}

In addition to these metrics, we also measure the Intersection over Union (IoU) of the building's outline. To do so, we convert the point clouds to 2D scatter plots by removing the height information and then find the outlines by using python's \texttt{alphashape} package. 

\subsection{Experiment Setup}
\label{sub:experiment_setup}
We augmented our dataset by padding the ground truth point cloud with additional points. To do so, we added a thin layer of points around the building outline. We set the height (i.e., z-value) of these points to a value lower than the median height of roof points offset by a fixed margin. We also ensured that the point density of the padding is similar to the building point density. We used this padded input as ground truth to train the model. Finally, we remove the padding from the predicted point cloud using a fixed threshold to extract the 3D building roof point cloud and use it to report our final results. 

Note that calculating the completeness, correctness, and quality metrics requires the extraction of roof planes from the ground truth and the predicted planes from the point cloud. To do so, we use the algorithm introduced by \cite{li2020roof} which identifies and segments the roof planes in the point cloud of lidar point cloud. We observe that the algorithm performs reasonably well on the ground truth lidar data and is able to extract the roof planes from the point cloud. We manually inspected the segmented results and found them to be satisfactory. Next, we use the same algorithm to extract the planar segments from the predicted point cloud to measure the performance of our model. 

To train our model, we divided our dataset into the train (85\%), validation (8\%), and test (7\%). Finally, set the value of $\alpha$ to 1 and use 3000 points in the random initial point cloud. We train the model five times and report the average across these multiple runs. \texttt{sat2pc} model is implemented in python language using the \texttt{pytorch} framework. We trained the model on an Nvidia RTX 3090 GPU. 
The \texttt{sat2pc} code, visualization tools, and the dataset will be made available for future research.

\section{Experimental Results}
\label{sec:experiments}

In this section, we thoroughly investigate our proposed approach and perform various baseline comparisons. We also perform ablation studies and report our insights.   

\begin{table*}[ht]
\footnotesize	
\begin{tabular}{c|c|c|c|c|c|c|c|c|c}
\toprule
\textbf{Model}                                                 & \textbf{\begin{tabular}[c]{@{}c@{}}Loss \\ Function\end{tabular}} & \textbf{\begin{tabular}[c]{@{}c@{}}EMD\\ With Pad\end{tabular}} & \textbf{\begin{tabular}[c]{@{}c@{}}Chamfer\\ With Pad\end{tabular}} & \textbf{\begin{tabular}[c]{@{}c@{}}EMD\\ Removed Pad\end{tabular}} & \textbf{\begin{tabular}[c]{@{}c@{}}Chamfer\\ Removed Pad\end{tabular}} & \textbf{\begin{tabular}[c]{@{}c@{}}Outline \\ IoU\end{tabular}} & \textbf{Completeness} & \textbf{Correctness} & \textbf{Quality} \\ \midrule
\textbf{\begin{tabular}[c]{@{}c@{}}GraphX~\cite{nguyen2019graphx}\end{tabular}} & \textbf{Chamfer}                                                  & 2.54                                                            & \textbf{0.08}                                                       & 1.08                                                               & \textbf{0.09}                                                          & \textbf{0.84}                                                   & 0.55                  & 0.65                 & 0.43             \\ \hline
\textbf{\begin{tabular}[c]{@{}c@{}}GraphX~\cite{nguyen2019graphx}\end{tabular}} & \textbf{EMD}                                                      & 0.29                                                            & 0.10                                                                & 0.25                                                               & 0.13                                                                   & 0.81                                                            & 0.47                  & 0.28                 & 0.21             \\ \hline
\textbf{PointSet\cite{fan2017point}}                                              & \textbf{Chamfer}                                                  & 2.43                                                            & 0.09                                                                & 0.87                                                               & 0.11                                                                   & 0.81                                                            & 0.32                  & \textbf{0.72}        & 0.30             \\ \hline
\textbf{sat2pc (Ours)}                                                & \textbf{\begin{tabular}[c]{@{}c@{}}Chamfer +\\ EMD\end{tabular}}  & \textbf{0.26}                                                   & \textbf{0.08}                                                       & \textbf{0.22}                                                      & 0.1                                                                    & \textbf{0.84}                                                   & \textbf{0.67}         & 0.67                 & \textbf{0.51}    \\ \bottomrule
\end{tabular}
\caption{Baseline comparisons of \texttt{sat2pc} with existing approaches. }
\label{tb:quantitative_results}
\end{table*}

\subsection{Qualitative Results}
\label{subsec:result_comparison}

We first compare our approach to prior work on 2D to 3D point cloud prediction. We consider two baseline techniques --- namely PointSetGeneration\cite{fan2017point} and GraphX-Conv~\cite{nguyen2019graphx}. Both techniques use a single image to predict the point cloud. To the best of our knowledge, there is no study on predicting building point cloud. Thus, these models were trained for our use case using the code released by the authors. In GraphX-Conv\cite{nguyen2019graphx}, the authors use only Chamfer loss to train their model. We modified the code to include EMD loss to train the GraphX models.  

Table\ref{tb:quantitative_results} shows the quantitative comparison of our approach with the baseline techniques. Our results show \texttt{sat2pc}, which combines both the Chamfer and EMD loss, outperforms in terms of quality and completeness metrics by a large margin. Recall that these metrics are calculated by extracting the roof planes from the point cloud and comparing them to the ground truth planar segments. This demonstrates the effectiveness of our \texttt{sat2pc} in reconstructing the geometric shapes in comparison to other baselines and shows that combining Chamfer and EMD loss helps in obtaining rich shape information.   

Although the PointSet technique has a higher correctness metric, the overall structure reconstructed is less accurate. In our analysis, we observed that the roof planar extraction technique could extract only a small number of planes from the PointSet output. Although these extracted planes had a high overlap with ground truth, indicating a higher correctness value, the technique could not extract many roof planar segments due to poor shape quality. In order to achieve good  shape quality, we need correctness, completeness, and quality value to be high. Our \texttt{sat2pc} model achieves high values in these metrics, indicating that the planar extraction technique was able to identify and segment roof planes from the output successfully.

We note that both GraphX-conv and \texttt{sat2pc} achieve similar IoU. This comes as no surprise as these models are capable of learning global semantics, and the augmentation of lidar points helps in achieving accurate outlines of the roof planes. Similarly, we observe that Chamfer and EMD loss are similar in most cases for GraphX-Conv and \texttt{sat2pc}. However, as indicated above, the output of GraphX-Conv performs lower compared to \texttt{sat2pc} with respect to shape information, indicating that Chamfer loss alone may not be sufficient in capturing the details. 

\textbf{Key takeaway:} Using Chamfer or EMD loss alone doesn't capture the characteristics of the building. In contrast, using a combination of Chamfer and EMD losses, \texttt{sat2pc} is able to obtain rich 3D shape information of roofs from 2D satellite images. \texttt{sat2pc} achieved 18.6\% improvement in quality metrics over the best baseline model.

\begin{figure*}[t]
\settoheight{\tempdima}{\includegraphics[width=.16\linewidth]{example-image-a}}%
\centering\begin{tabular}{@{}c@{ }c@{ }c@{ }c@{ }c@{ }c@{}}
&\textbf{Input image} & \textbf{PointSet} & \textbf{GraphX-Conv} & \textbf{Sat2PC} & \textbf{Ground truth}\\
\rowname{Sample A}&
\includegraphics[width = .12\linewidth]{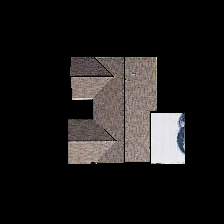} &
\includegraphics[width =.12\linewidth]{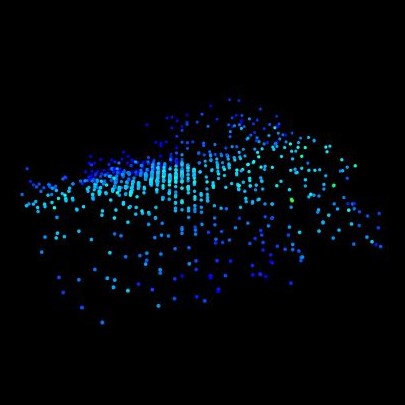} &
\includegraphics[width = .12\linewidth]{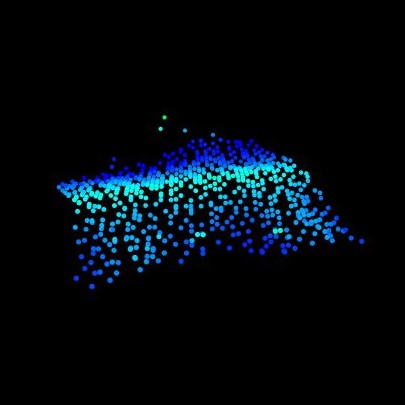} &
\includegraphics[width = .12\linewidth]{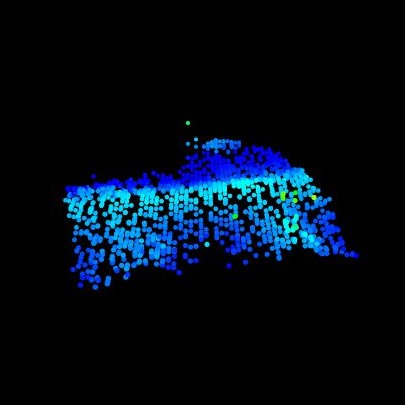} &
\includegraphics[width = .12\linewidth]{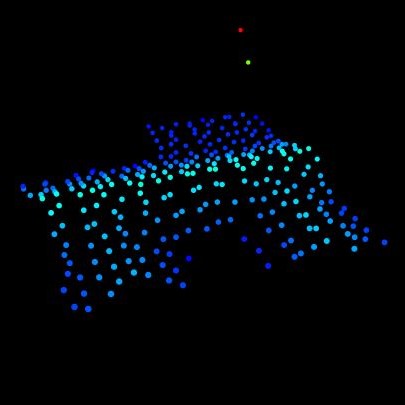} \\

\rowname{Sample B}&
\includegraphics[width = .12\linewidth]{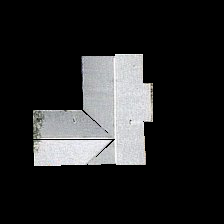} &
\includegraphics[width =.12\linewidth]{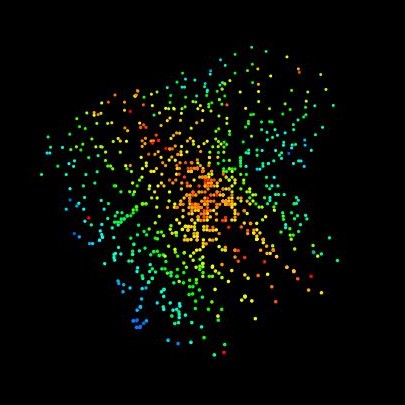} &
\includegraphics[width = .12\linewidth]{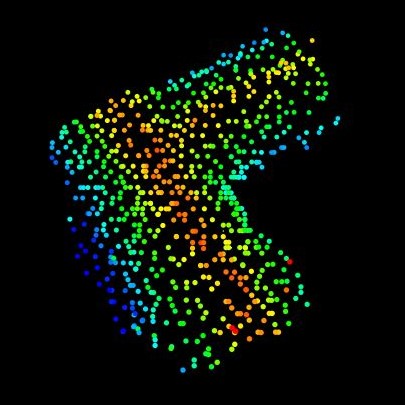} &
\includegraphics[width = .12\linewidth]{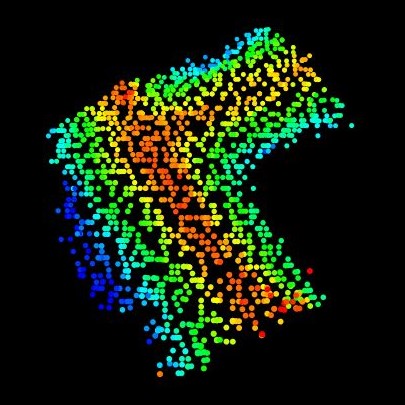} &
\includegraphics[width = .12\linewidth]{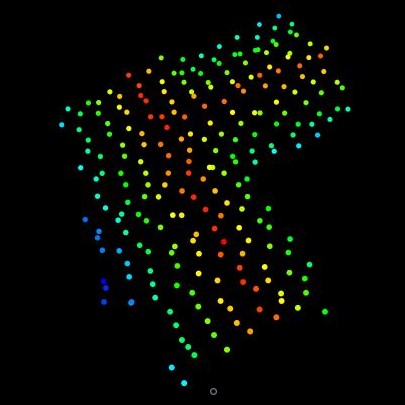} \\

\rowname{Sample C}&
\includegraphics[width = .12\linewidth]{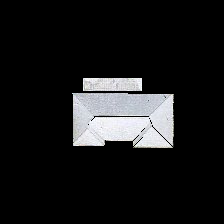} &
\includegraphics[width =.12\linewidth]{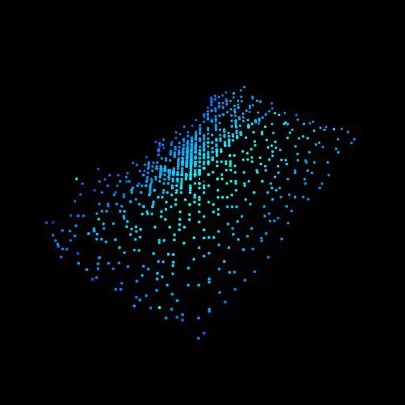} &
\includegraphics[width = .12\linewidth]{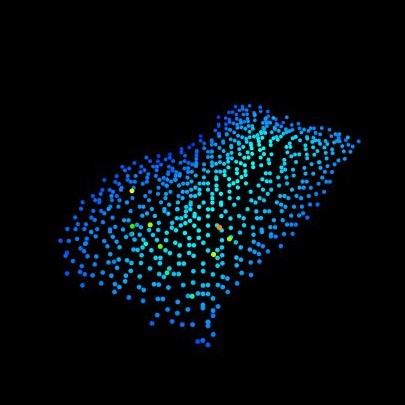} &
\includegraphics[width = .12\linewidth]{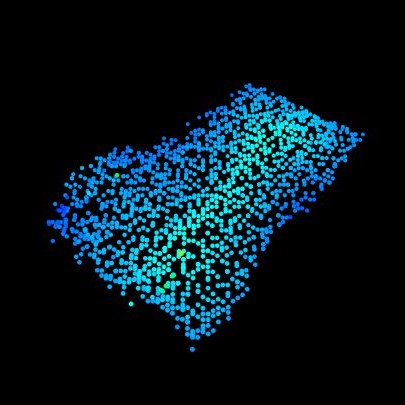} &
\includegraphics[width = .12\linewidth]{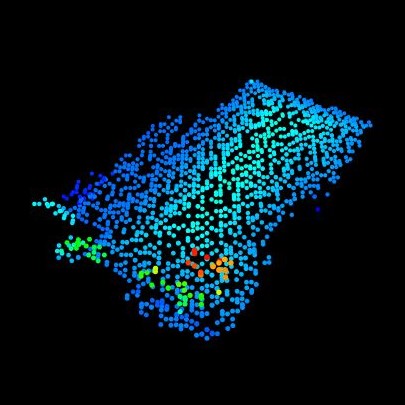} \\

\rowname{Sample D}&
\includegraphics[width = .12\linewidth]{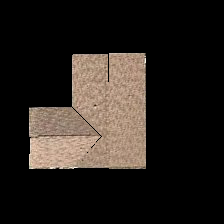} &
\includegraphics[width =.12\linewidth]{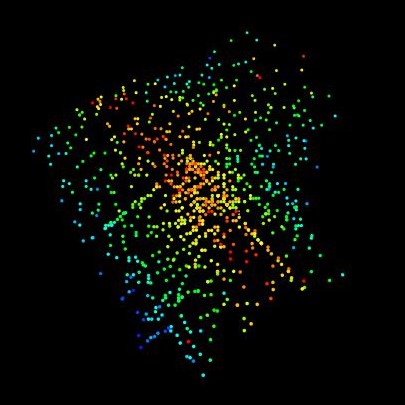} &
\includegraphics[width = .12\linewidth]{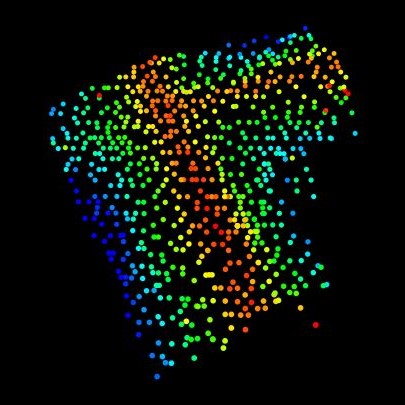}&
\includegraphics[width = .12\linewidth]{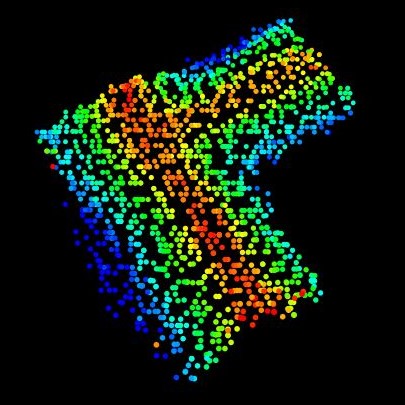} &
\includegraphics[width = .12\linewidth]{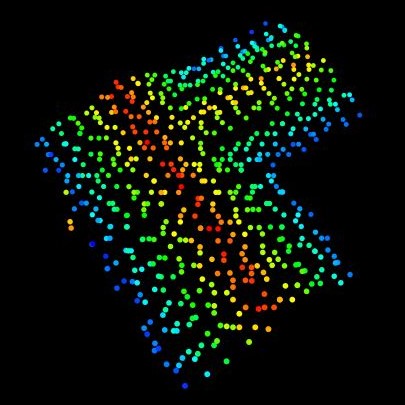}\\
\end{tabular}

\caption{Qualitative results of \texttt{sat2pc}, GraphX-Conv, and PointSet. The color of each point in the point clouds indicates its height relative to the points in that point cloud.}

\label{fig:qualitative}
\end{figure*}

\begin{figure}[t]
\settoheight{\tempdima}{\includegraphics[width=.31\linewidth]{example-image-a}}%
\centering\begin{tabular}{@{}c@{ }c@{ }c@{ }c@{}}
&\textbf{GraphX-Conv} & \textbf{Sat2PC} & \textbf{Ground truth}\\
\rowname{Sample A}&
\includegraphics[width = .23\linewidth]{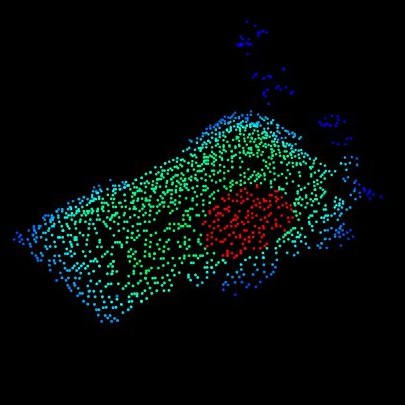} &
\includegraphics[width =.23\linewidth]{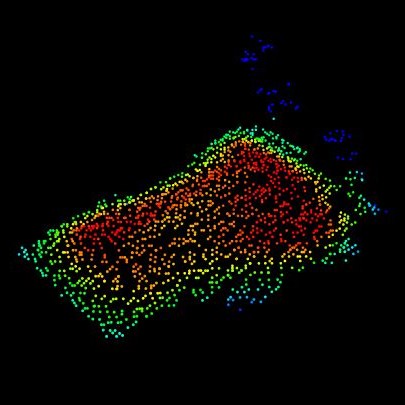} &
\includegraphics[width = .23\linewidth]{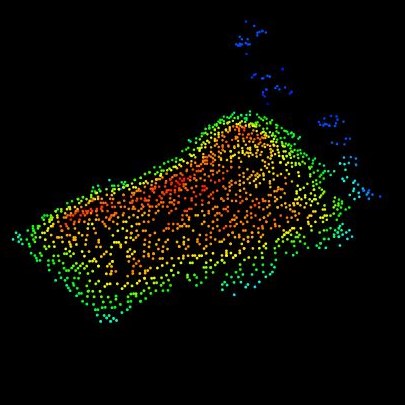}\\

\rowname{Sample B}&
\includegraphics[width = .23\linewidth]{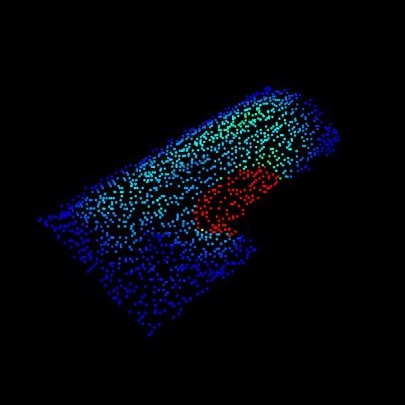} &
\includegraphics[width =.23\linewidth]{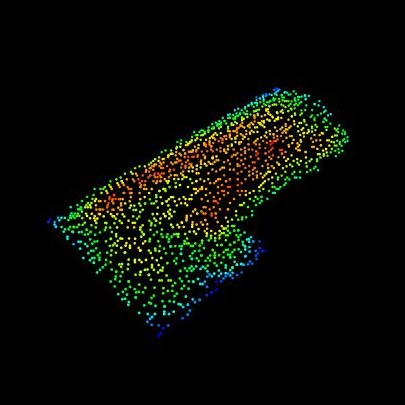} &
\includegraphics[width = .23\linewidth]{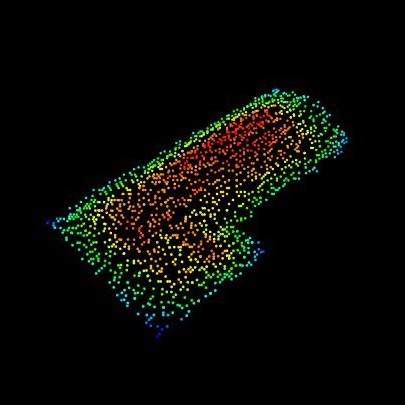}\\

\rowname{Sample C}&
\includegraphics[width = .23\linewidth]{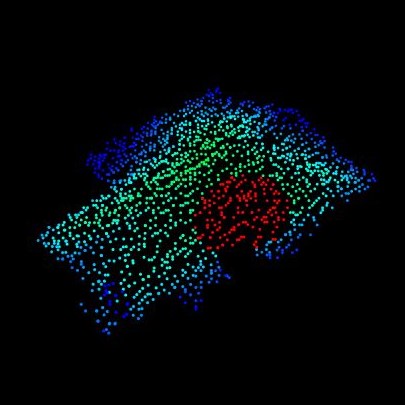} &
\includegraphics[width =.23\linewidth]{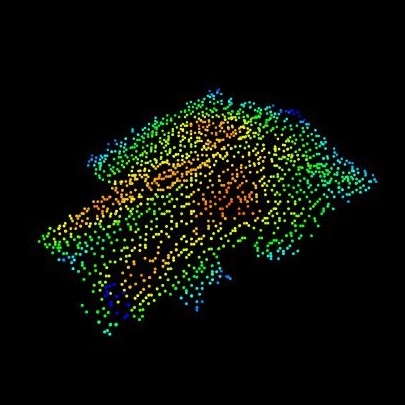} &
\includegraphics[width = .23\linewidth]{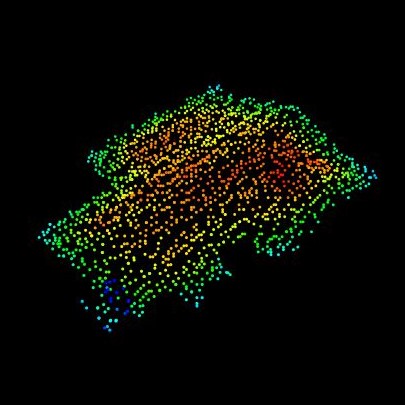}\\

\end{tabular}

\caption{Density of predicted points overlaid on ground truth point cloud. Sat2PC generates a  similar density to ground truth density. The colors indicate the density of each points where red shows high density and blue shows low density.} 
\label{fig:density}
\end{figure}

\begin{figure}[t]
\includegraphics[width=.75\linewidth]{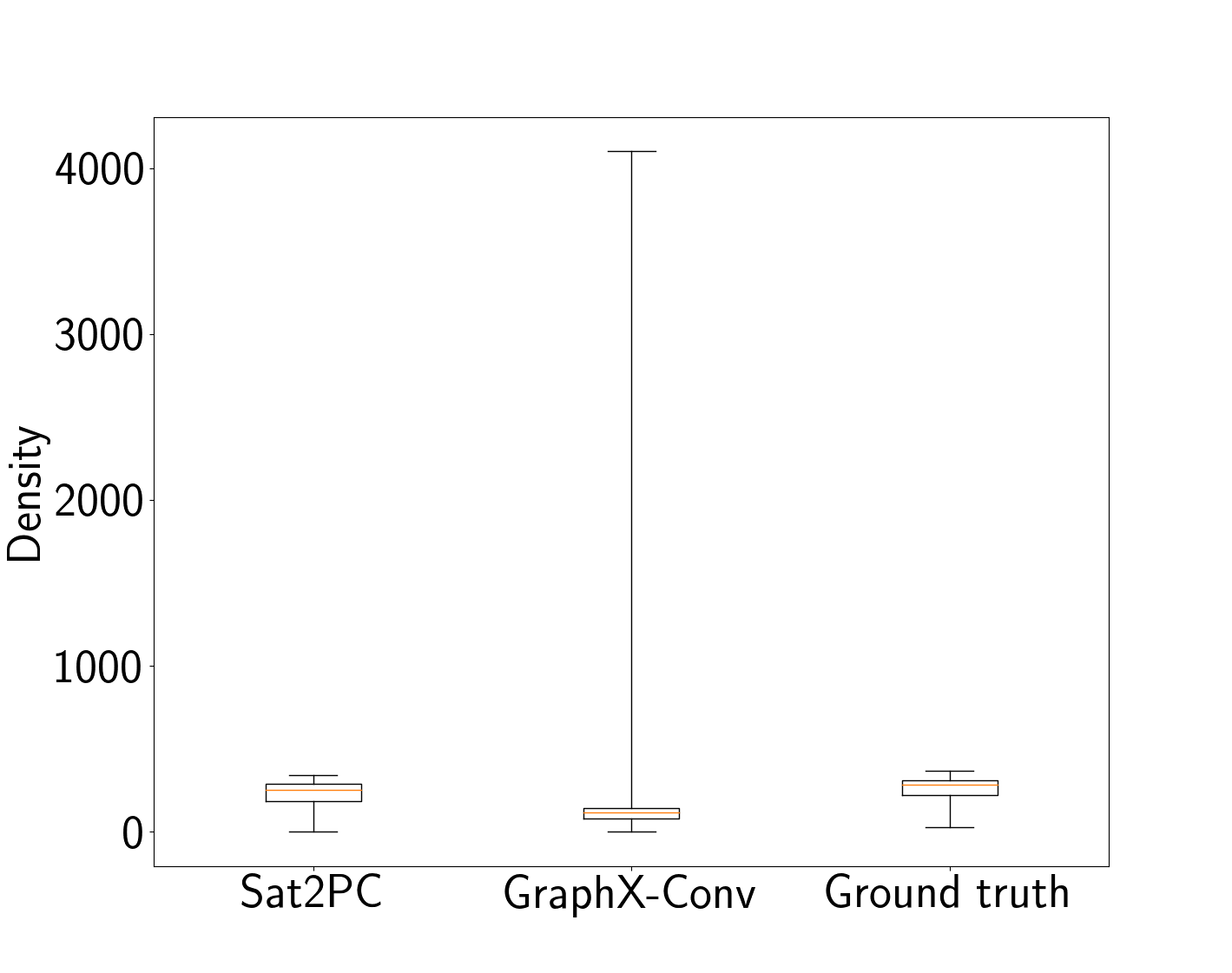}
  \caption{Density of areas around each point from ground truth point cloud in the predicted point clouds.}
  \label{fig:density_box}
\end{figure}

\subsection{Qualitative Results}
Figure \ref{fig:qualitative} shows the qualitative results. As depicted in the figure, our approach helps produce better 3D point clouds of buildings. In particular, we observe that PointSet technique is unable to yield rich shape information when compared to GraphX and \texttt{sat2pc}. Although GraphX-Conv predicts the overall shape, we observe that it misses details that are otherwise observed in \texttt{sat2pc} shape output. For example, in sample A, the shape generated by GraphX-Conv fails to capture the gap between two adjoining roofs and identifies it as a single roof plane. In contrast, \texttt{sat2pc} output predicts the gap and preserves the overall structure. This problem is also depicted in sample C, where GraphX-Conv fails to capture the gap. 

We further investigate the density of predicted points. To do so, we use the ground truth points and calculate the number of points within a fixed radius in the GraphX and \texttt{sat2pc} point cloud. Figure~\ref{fig:density} shows the density of points compared to the ground truth lidar data. We observe that GraphX with Chamfer loss cluster points into high-density patches in regions. This is demonstrated by the red patches, which indicate a higher density of points compared to the ground truth. In contrast, \texttt{sat2pc} the points are not clustered and uniformly spread across, resulting in a similar density as the ground truth data. We also calculate the number of points within a fixed radius. As shown in Figure~\ref{fig:density_box}, GraphX has high variance, indicating that most points are clustered within a region. In contrast, the density distribution of  \texttt{sat2pc} is comparable to the ground truth data.  We also observe that the points are not uniform predicted by GraphX, with significant gaps between neighboring points. On the other hand, \texttt{sat2pc} predicts smooth points and has a much better appearance. 

\textbf{Key takeaway:} \texttt{sat2pc} provides a uniform point cloud and has a much better appearance in comparison to other baseline techniques. The ability to combine the density-awareness of EMD loss and the global structure of chamfer loss results in a rich point cloud representation.

\label{sec:ablation}

\begin{table*}[t]
\footnotesize	
\begin{tabular}{c|c|c|c|c|c|c|c|c|c}
\toprule
\textbf{\begin{tabular}[c]{@{}c@{}}Deformation\\ Loss\end{tabular}} & \textbf{\begin{tabular}[c]{@{}c@{}}Refinement\\ Loss\end{tabular}} & \textbf{\begin{tabular}[c]{@{}c@{}}EMD\\ With Pad\end{tabular}} & \textbf{\begin{tabular}[c]{@{}c@{}}Chamfer\\ With Pad\end{tabular}} & \textbf{\begin{tabular}[c]{@{}c@{}}EMD\\ Removed Pad\end{tabular}} & \textbf{\begin{tabular}[c]{@{}c@{}}Chamfer\\ Removed Pad\end{tabular}} & \textbf{\begin{tabular}[c]{@{}c@{}}Outline \\ IoU\end{tabular}} & \textbf{Completeness} & \textbf{Correctness} & \textbf{Quality} \\ \midrule
\textbf{Chamfer}                                                    & \textbf{EMD}                                                       & 0.26                                                            & \textbf{0.08}                                                       & \textbf{0.22}                                                      & \textbf{0.1}                                                           & \textbf{0.84}                                                   & \textbf{0.67}         & 0.67                 & \textbf{0.51}    \\ \hline
\textbf{Chamfer}                                                    & \textbf{Chamfer}                                                   & 3.13                                                            & 0.1                                                                 & 1.18                                                               & 0.13                                                                   & 0.8                                                             & 0.44                  & \textbf{0.73}        & 0.39             \\ \hline
\textbf{EMD}                                                        & \textbf{EMD}                                                       & 0.26                                                            & \textbf{0.08}                                                       & \textbf{0.22}                                                      & 0.12                                                                   & \textbf{0.84}                                                   & 0.65                  & 0.62                 & 0.47             \\ \hline
\textbf{EMD}                                                        & \textbf{Chamfer}                                                   & \textbf{0.25}                                                   & \textbf{0.08}                                                       & \textbf{0.22}                                                      & 0.11                                                                   & \textbf{0.84}                                                   & 0.64                  & 0.61                 & 0.45             \\ \bottomrule
\end{tabular}
\caption{
Analyzing the impact of changing the order of loss functions in different modules in our architecture.
}
\label{tb:effect_loss}
\end{table*}

\subsection{Effect of Loss Function}
We analyze the impact of different Chamfer and EMD loss combinations within the \texttt{sat2pc} architecture. Table~\ref{tb:effect_loss} shows the overall performance for different combinations, with $\alpha$ set to 1. We observe that the overall quality drops when we use the same loss function in \texttt{sat2pc}. This is because Chamfer and EMD loss captures different shape characteristics: global structure versus local density. We also observe that the order of loss function in the architecture can significantly impact the final output. In particular, using the Chamfer loss function on the point cloud deformation module and EMD on the refinement module offers the best performance. Since the point cloud deformation module has access to the input image, we conjecture that having Chamfer loss helps capture the overall details from that image.
On the other hand, density and small refinements based on the point's location can be effective without having direct access to the image features. Hence, using EMD as a density-aware loss on the refinement module's output can help achieve a uniform and smooth shape. 
When we reverse the order of the loss function,  the overall quality drops --- potentially due to its inability to capture more details from the image.

\textbf{Key takeaway:} 
The combination of different losses helps in improving the overall performance of the model. Using Chamfer with access to image details and EMD at the refinement stage can boost the model's performance.

\subsection{Effect of Alpha}

\begin{table*}[t]
\footnotesize
\begin{tabular}{c|c|c|c|c|c|c|c|c}
\toprule
\textbf{$\alpha$} & \textbf{\begin{tabular}[c]{@{}c@{}}EMD\\ W/Pad\end{tabular}} & \textbf{\begin{tabular}[c]{@{}c@{}}Chamfer\\ W/Pad\end{tabular}} & \textbf{\begin{tabular}[c]{@{}c@{}}EMD\\ No Pad\end{tabular}} & \textbf{\begin{tabular}[c]{@{}c@{}}Chamfer\\ No Pad\end{tabular}} & \textbf{\begin{tabular}[c]{@{}c@{}}Outline \\ IoU\end{tabular}} & \textbf{Completeness} & \textbf{Correctness} & \textbf{Quality} \\ \midrule
\textbf{0.2}   & 0.37                                                         & 0.17                                                             & 0.23                                                          & \textbf{0.09}                                                     & 0.83                                                            & 0.6                   & 0.64                 & 0.44             \\ \hline
\textbf{0.3}   & 0.34                                                         & 0.15                                                             & 0.23                                                          & 0.1                                                               & 0.83                                                            & 0.61                  & 0.63                 & 0.45             \\ \hline
\textbf{0.5}   & 0.29                                                         & 0.1                                                              & \textbf{0.22}                                                 & 0.1                                                               & \textbf{0.84}                                                   & 0.64                  & 0.61                 & 0.45             \\ \hline
\textbf{0.8}   & 0.28                                                         & 0.09                                                             & \textbf{0.22}                                                 & 0.1                                                               & \textbf{0.84}                                                   & 0.63                  & \textbf{0.67}        & 0.48             \\ \hline
\textbf{1}     & 0.26                                                         & 0.08                                                             & \textbf{0.22}                                                 & 0.1                                                               & \textbf{0.84}                                                   & \textbf{0.67}         & \textbf{0.67}        & \textbf{0.51}    \\ \hline
\textbf{1.5}   & \textbf{0.25}                                                & \textbf{0.07}                                                    & \textbf{0.22}                                                 & 0.1                                                               & \textbf{0.84}                                                   & 0.66                  & 0.66                 & 0.49             \\ \hline
\textbf{2}     & \textbf{0.25}                                                & \textbf{0.07}                                                    & 0.23                                                          & 0.1                                                               & \textbf{0.84}                                                   & 0.65                  & 0.62                 & 0.47             \\ \bottomrule
\end{tabular}
\caption{Impact of varying alpha value on performance.}
\label{tb:alpha_effect}
\end{table*}

Table \ref{tb:alpha_effect} shows the different $\alpha$ values on the overall performance of \texttt{sat2pc}. Note that the $\alpha$ parameter controls the overall importance of each loss function in the architecture. 
We observe that the overall quality of output increases as we increase from $0.2$ to $1$. When $\alpha=1$, both Chamfer and EMD loss have equal weightage, and we observe better output quality. However, when the $\alpha$ value is greater than 1, the model's performance reduces in terms of quality, correctness, and completeness metrics. 

\textbf{Key takeaway:} When $\alpha=1$, the trained \texttt{sat2pc} 
models perform best in terms of completeness, correctness, and quality metrics --- likely due to better-predicted shape.

\subsection{Effect of Refinement Module}
\begin{table}[]
\footnotesize
\begin{tabular}{c|c|c|c}
\toprule
\textbf{Point Cloud Output}  & \textbf{Completeness} & \textbf{Correctness} & \textbf{Quality} \\ \midrule
\textbf{Point Cloud Deformation} & 0.65                  & 0.64                 & 0.49             \\ \hline
\textbf{PCD + Refinement}  & \textbf{0.67 }                 & \textbf{0.67 }                & \textbf{0.51  }           \\ \bottomrule
\end{tabular}
\caption{Impact of refinement module on performance.}
\label{tb:effect_refine}
\end{table}

Next, we analyze the output of the Point Cloud Deformation (PCD) module and refinement module in \texttt{sat2pc}. Recall that PCD module in \texttt{sat2pc} produces an intermediate point cloud, which is further processed by the refinement module.  
To do so, we ran the planar extraction algorithm on the output of the point cloud deformation(PCD) module and refinement module and calculated the completeness, correctness, and quality. Table \ref{tb:effect_refine} shows  the  performance of the point cloud output. We observe that the final output from the refinement module further refines and improves the 3D shape. This is demonstrated by the higher completeness, correctness, and quality values. 

\textbf{Key takeaway:} 
The refinement module in \texttt{sat2pc} helps improve the overall shape of the final point cloud. 

\subsection{Effect of Point Cloud Resolution}

\begin{table*}[]
\footnotesize
\begin{tabular}{c|c|c|c|c|c|c|c|c|c}
\toprule
\textbf{\begin{tabular}[c]{@{}c@{}}Number \\ of Points\end{tabular}} & \textbf{\begin{tabular}[c]{@{}c@{}}Time Per\\ Epoch(s)\end{tabular}} & \textbf{\begin{tabular}[c]{@{}c@{}}EMD\\ With Pad\end{tabular}} & \textbf{\begin{tabular}[c]{@{}c@{}}Chamfer\\ With Pad\end{tabular}} & \textbf{\begin{tabular}[c]{@{}c@{}}EMD\\ Removed Pad\end{tabular}} & \textbf{\begin{tabular}[c]{@{}c@{}}Chamfer\\ Removed Pad\end{tabular}} & \textbf{\begin{tabular}[c]{@{}c@{}}Outline \\ IoU\end{tabular}} & \textbf{Completeness} & \textbf{Correctness} & \textbf{Quality} \\ \midrule
\textbf{3000}                                                        & 31.5                                                                 & 0.26                                                            & \textbf{0.08}                                                       & 0.22                                                               & \textbf{0.1}                                                           & 0.84                                                            & 0.67                  & 0.67                 & 0.51             \\ \hline
\textbf{4000}                                                        & 37.8                                                                 & \textbf{0.17}                                                   & \textbf{0.08}                                                       & \textbf{0.17}                                                      & \textbf{0.1}                                                           & \textbf{0.85}                                                   & 0.68                  & \textbf{0.7}         & \textbf{0.52}    \\ \hline
\textbf{6000}                                                        & 51.7                                                                 & 0.29                                                            & \textbf{0.08}                                                       & 0.21                                                               & \textbf{0.1}                                                           & \textbf{0.85}                                                   & 0.66                  & 0.6                  & 0.45             \\ \hline
\textbf{8000}                                                        & 62                                                                   & 0.21                                                            & \textbf{0.08}                                                       & 0.22                                                               & \textbf{0.1}                                                           & \textbf{0.85}                                                   & \textbf{0.69}         & 0.61                 & 0.47             \\ \bottomrule
\end{tabular}
\caption{Performance of \texttt{sat2pc} in predicting point cloud output with different point cloud size.}
\label{tb:effect_resolution}
\end{table*}
Finally, we vary the initial random cloud size to analyze the impact of point cloud size on the overall shape quality. Table \ref{tb:effect_resolution} shows the results for  varying point cloud size. We observe that increasing the number of points from 3000 to 4000 improves the overall results, although the improvement is marginal.
We also visually inspected the point cloud, and the predicted point surface was smooth and uniform. However, further increasing the point cloud size reduces the overall performance. In particular, we see a reduction in quality and correctness metric --- presumably due to the resolution of ground truth lidar data, which has a maximum of 3000 points. In terms of training time, increasing the number of points in the point cloud also increases training time. In particular, increasing the number of points from 3000 to 4000 almost doubles the training time. 

\textbf{Key Takeaway:} Increasing the number of points to 4000 improves the overall results.

However, a further increase reduces the overall quality of the output.

\section{Discussion}
\label{sec:discussion}
Our current approach predicts point cloud at an individual building level. However, it is possible to easily scale our approach to create 3D urban models using satellite images at scale cost-effectively. Recall that our approach requires a building outline. This can easily be obtained using online mapping APIs (e.g., OpenStreet Maps). Moreover, we can use existing vision-based models to extract building roofs with high accuracy. We can then use our model to predict the building roof point clouds using the segmented 2D roof and use the location coordinates to project these point clouds. There are also other design considerations for scaling our proposed approach. First, our approach did not consider occluded roofs from trees or nearby structures. While our work can predict point cloud of roof images with shadows from nearby trees present in our dataset, partially occluded images will require hallucination of the roof size and shape. We believe that architectures such as Generative  Adversarial  Networks(GAN) can help recover partially occluded roofs and remain part of future work. To generate an urban model will also require the prediction of trees and other structures (e.g., utility poles). We observed that, in some cases, the final point cloud also detected the presence of trees, marked by the higher z-values of points along the edges. There has been recent work on detecting and predicting tree crowns and their height using images\cite{weinstein2019individual}. We believe that we can leverage these approaches to predict trees and vegetation. Separately, urban models will require predicting the height of the building. And, there has been recent work on predicting building height using single 2D images~\cite{ghamisi2018img2dsm, srivastava2017joint, liu2020im2elevation}. 

\section{Related Work}
\label{sec:relatedwork}

\textbf{2D to 3D Object Reconstruction} Prior work on 3D reconstruction relied primarily on stereo or multi-view images\cite{ kar2017learning, wen2019pixel2mesh++}. However, the recent success of deep learning techniques in 3D reconstruction has seen renewed interest in constructing models from single images. Although an ill-posed problem, many techniques have attempted to predict the 3D shape from a single image using prior knowledge and learning the geometries available in the image. 3D shape of an object can be represented in various forms, namely --- voxel\cite{liu2021voxel}, mesh\cite{wen2019pixel2mesh++}, and point cloud~\cite{nguyen2019graphx, fan2017point}. While there have been several works on predicting these different 3D representations, our work primarily focuses on point cloud representation because of the accessibility of point cloud building data. In addition, point cloud representations are also computationally more efficient compared to voxel-based representations\cite{tatarchenko2017octree}. Fan et al.~\cite{fan2017point} introduced an architecture that directly generates point cloud representations using single images. Subsequently, there have been follow-up work in this direction\cite{nguyen2019graphx}. Unlike prior work, we use a combination of loss functions to show significant improvement in improving the quality of the point cloud. Further, we evaluate our work on real-world lidar sensor data of buildings prone to be noisy.   

There has also been considerable work on network architectures that complete or upsamples 3D shape\cite{wang2020cascaded}. These upsampling techniques generate a dense and uniform point cloud from sparse point cloud data. Similarly, work on 3D shape completion completes missing parts of point cloud due to occlusions\cite{liu2020morphing, wang2020cascaded}. As in \cite{wang2020cascaded}, these techniques may also use upsampling and refinement to ensure that the final 3D shape does not deviate from the overall structure. We use the residual network in our work to help the model learn the density. 

\textbf{3D Building reconstruction.}

There have been several decades of research in developing large-scale urban models. Generating 3D building shapes from images is a classical remote sensing problem, as satellite images provide a cost-effective means of collecting large-scale building data. For example, several techniques use multi-view images to extract the shape or use 3D primitives to estimate the shape of the building\cite{li20213d, alidoost20192d, mahmud2020boundary}. With the availability of point cloud lidar, studies have focused on combining lidar and image features to create accurate 3D urban models\cite{ledoux20213dfier}. Recently, deep learning-based approaches have been proposed to estimate the height of a building from single 2D images\cite{liu2020im2elevation}. Separately, there has been work on segmenting images to estimate the 3D shape of the roof. In \cite{alidoost20192d}, the authors propose a multi-scale convolutional–deconvolutional network (MSCDN) that predicts the Digital Surface Model (DSM) of the buildings from a single aerial RGB image. In contrast, our work focuses on predicting a point cloud from 2D images. To the best of our knowledge, this is the first work that aims to predict point cloud from a single satellite image.

\section{Conclusion}
In this paper, we present \texttt{sat2pc}, an architecture for predicting building 3D point cloud from a single satellite image. In addition, we provide a new building-oriented dataset that includes high-resolution satellite images and LiDAR information along with building outlines. 
We extensively evaluated our approach and performed ablation studies to show the efficacy of our proposed approach. Our results showed that \texttt{sat2pc} can reconstruct the 3D point cloud from 2D images. Furthermore, \texttt{sat2pc} outperforms existing baseline techniques in terms of the overall shape quality of 3D models. We also showed that planar extraction algorithms could extract planar segments from the predicted point cloud. 

\bibliographystyle{ACM-Reference-Format}

\raggedright
\bibliography{paper}

\end{document}